\documentclass{article}
\usepackage{iclr2027_conference,times}
\usepackage[T1]{fontenc}
\usepackage[utf8]{inputenc}
\usepackage{amsmath,amssymb,amsthm}
\usepackage{graphicx,booktabs,tabularx,multirow}
\usepackage{algorithm,algpseudocode}
\usepackage{xcolor}
\usepackage{caption}
\usepackage{xspace}
\usepackage{makecell}
\usepackage[hidelinks]{hyperref}
\usepackage{url}
\newcommand{\method}{ProactiveVLA\xspace}

\title{ProactiveVLA: Augmenting Embodied Memory through Proactive Environment Exploration}

\iclrfinalcopy % 开启显示作者信息
\makeatletter
\def\@oddhead{}
\def\@evenhead{}
\makeatother

\author{Shizuo Tian$^{1}\thanks{\ \ Indicates equal contribution.}$\hspace{0.5em}, Haodong Luo$^{1*}$, Yutong Li$^{1}$, Yuebing Song$^{2}$, Yunxin Liu$^{1}$, Yuanchun Li$^{1}$\thanks{\ \ Corresponding Author: Yuanchun Li (\texttt{liyuanchun@air.tsinghua.edu.cn}).}\hspace{0.5em}\\
$^{1}$Tsinghua University,
$^{2}$Tongji University
}

\begin{document}
\maketitle
\begin{abstract}
Rapid adaptation to a new environment requires a robot to acquire useful knowledge about local objects, states, and interactions from limited experience.
Systems that combine a reasoning agent with a frozen vision-language-action model (VLA) can adapt through execution feedback and memory, making the choice of experience central to their effectiveness.
Repeated practice of a target task may refine a familiar solution while leaving other interactions relevant to changed conditions untested.
We introduce ProactiveVLA, which uses proactive environment exploration to acquire reusable knowledge for deployment-time adaptation.
After completing an initial task, the agent allocates the remaining interaction budget to self-proposed goals covering object affordances, state-changing interactions, and compositions of interactions.
It verifies execution outcomes and consolidates both task-directed and exploratory experience into memory that guides subsequent planning and control.
ProactiveVLA outperforms the baselines under the same turn budget on LIBERO-Pro and RoboCasa365 Composite-Seen. On LIBERO-Pro Goal-T, with at most one VLA primitive invocation allowed during evaluation, ProactiveVLA completes 48\% of instances, compared with 19\% for the state-of-the-art task-refinement baseline.
\end{abstract}

\section{Introduction}
\label{sec:introduction}
A robot deployed in an unfamiliar environment must use its existing capabilities with new objects, layouts, and task requirements.
Vision-language-action models (VLAs), including OpenVLA and $\pi_{0.5}$, provide pretrained manipulation capabilities learned from broad robot data \citep{kim2024openvla,physicalintelligence2025pi05}.
However, possessing these capabilities does not establish how they work in a particular deployment environment.
The robot needs experience with local object uses, action outcomes, and interaction sequences to guide its decisions.
Grounding decisions in available skills and environment feedback helps connect language-level knowledge to feasible robot behavior \citep{ichter2023saycan,huang2023innermonologue}.
With limited opportunities for interaction, rapid adaptation depends on acquiring information that supports task execution beyond the specific conditions already encountered.

Recent advances in agentic robotics use language models to organize perception, planning, and control through executable robot interfaces~\citep{liang2022codeaspolicies,singh2023progprompt,fu2026capx,zhang2026harnessvla}.
Such systems make experience acquisition an explicit design choice, since the agent determines which interactions to attempt and what knowledge to preserve.
Harness VLA provides a state-of-the-art agentic framework for adapting frozen VLAs, exposing the VLA as a callable manipulation primitive and combining it with analytic control and execution memory.
During its exploratory bootstrapping phase, Harness VLA searches for a working solution on a reference task instance and refines its execution memory across attempts \citep{zhang2026harnessvla}.
We refer to experience acquisition that keeps the provided task objective fixed while improving its execution as \emph{task refinement}.
Although this can improve a working solution, the resulting experience may remain concentrated around the same task requirements.
For example, practicing placement with an unobstructed target can refine grasping and release but may provide little direct evidence about clearing space when the target becomes obstructed.
Handling this change may require an interaction absent from the successful trajectory.
Improving a familiar solution and expanding the conditions under which the robot can act are therefore different uses of the same interaction budget.

Self-directed play already provides a route to acquiring reusable robot skills, as demonstrated by RATS \citep{zhang2026playful}.
For rapid adaptation of a VLA-based agentic system, we focus on using limited deployment-time interaction to acquire knowledge relevant to the local environment and its tasks.
This knowledge concerns the roles objects can serve, the changes an interaction can produce, and the ways interactions can be combined.
Under a limited interaction budget, actively acquiring diverse and reusable environment knowledge can improve adaptation to changes in tasks and execution conditions beyond practicing a single solution.

We introduce \method{}, an adaptation framework that enables agentic VLA systems to proactively acquire and organize environment knowledge through interaction.
The agent first tries the target task to obtain concrete experience with the local environment and controller.
It then uses the remaining budget to propose and execute goals at three levels, covering object affordances, state-changing interactions, and interaction compositions.
Visual outcome checks connect each proposal to observed execution evidence.
A memory-management process organizes experience within scenes, synthesizes candidate knowledge across scenes, and uses an independent critic to review proposed rules before they guide subsequent execution.
Both initial task experience and exploration outcomes contribute to this memory.
Adaptation takes place through the agent's acquired knowledge while the VLA remains frozen. The experiments show that ProactiveVLA outperforms Harness VLA under the same adaptation-turn budget on LIBERO-Pro. Under the same limited VLA primitive invocation budget, ProactiveVLA achieves higher task success, with an advantage when only a small number of VLA calls are available.

Our contributions are threefold.
First, we study deployment-time adaptation as the acquisition of reusable environment knowledge under a fixed interaction budget.
Second, we develop a task-guided proactive exploration method that couples three levels of goal generation with outcome verification and memory consolidation.
Third, we evaluate \method{} on LIBERO-Pro and RoboCasa365 Composite-Seen, covering controlled task and spatial variations as well as multi-step household tasks, and analyze how exploration structure and memory consolidation affect adaptation.

% Related Work for ProactiveVLA
% Suggested citation keys are listed at the end of this file.

\section{Related Work}

\paragraph{Vision-language-action models and generalization.}
Vision-language-action models learn language-conditioned visuomotor policies from large collections of robot trajectories and vision-language data.
RT-2 introduced the VLA formulation by jointly representing language and robot actions, while Open X-Embodiment expanded cross-embodiment robot data at scale \citep{brohan2023rt2, openx2023}.
OpenVLA provided an open generalist VLA, and subsequent models such as $\pi_0$ and $\pi_{0.5}$ scaled continuous action generation and heterogeneous robot training \citep{kim2024openvla, black2024pi0, physicalintelligence2025pi05}.
These models obtain strong manipulation capabilities from pretraining, but robustness can decrease when deployment changes task bindings, object arrangements, or compositional requirements.
LIBERO-Pro explicitly evaluates such changes through controlled perturbations of LIBERO tasks \citep{liu2023libero, zhou2025liberopro}, while RoboCasa365 provides diverse multi-step household manipulation tasks in kitchen environments \citep{nasiriany2026robocasa365}.

\paragraph{Agentic adaptation around robot policies.}
Language-model agents provide another route to generalization by composing perception and control interfaces around pretrained robot capabilities. 
Language-model agents interleave reasoning with actions and environment feedback \citep{yao2023react}.
In robotics, this interaction connects high-level decisions to perception and control interfaces.
Code as Policies and ProgPrompt use executable programs or program-like plans to connect language-model reasoning with robot APIs \citep{liang2022codeaspolicies, singh2023progprompt}.
More recent systems increasingly treat the policy as one component inside a persistent execution harness.
Harness VLA exposes a frozen VLA as a contact-rich manipulation primitive and combines it with high-level agent reasoning, analytic control, and execution memory \citep{zhang2026harnessvla}.
Recent agentic systems further explore self-improvement through interaction feedback, skill discovery, runtime recovery, and experience accumulation \citep{lu2026aspire,xiao2026enpire,zhang2026playful,ding2026zetta}.
\method{} studies adaptation efficiency under a limited interaction budget, focusing on what experience should be acquired from a new environment.
It allocates this budget to acquiring reusable environment knowledge rather than concentrating additional experience on a single task.

\paragraph{Autonomous exploration and reusable experience.}
Prior work has also studied how agents can autonomously acquire experience beyond a single externally specified task.
Plan2Explore uses task-agnostic exploration to learn a world model for later adaptation \citep{sekar2020plan2explore}.
ELLM uses language models to propose semantically meaningful exploration goals, showing that language-guided exploration can improve the usefulness of pretraining experience \citep{du2023ellm}.
Voyager combines automatic curriculum generation, environment interaction, and a persistent skill library for open-ended embodied learning \citep{wang2023voyager}.
More closely related to robot manipulation, recent agentic systems use self-directed interaction to discover reusable skills or behavioral knowledge that can later support downstream tasks \citep{lu2026aspire, zhang2026playful}.
Experience can also be consolidated through reflection and verification.
Reflexion retains verbal reflections in episodic memory, while ReasoningBank distills reasoning strategies from self-judged successful and failed attempts and updates memory across tasks \citep{shinn2023reflexion,ouyang2026reasoningbank}.
REFLECT summarizes multimodal robot experience to diagnose failures and guide subsequent correction \citep{liu2023reflect}.
ProactiveVLA shares the premise that interaction should create reusable knowledge, but studies a budgeted preparation problem.

\section{Methodology}
\label{sec:method}
\begin{figure}[t]
\centering
\includegraphics[width=\linewidth]{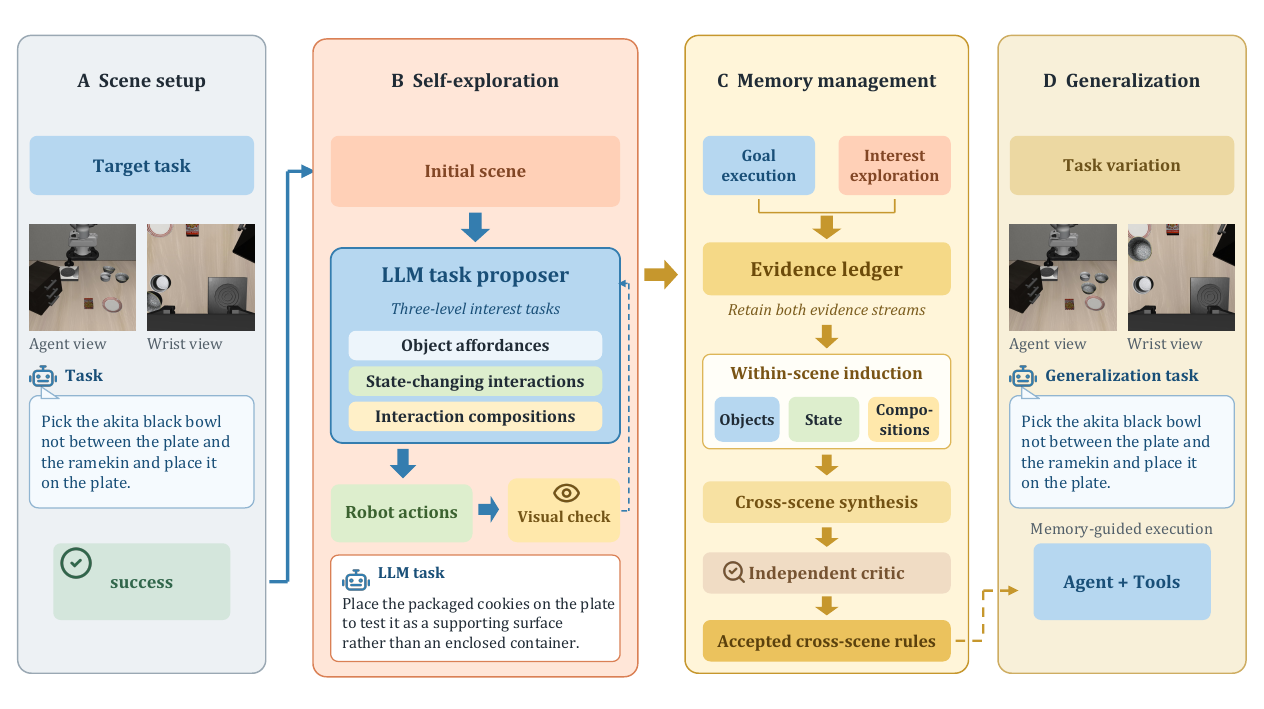}
\caption{Overview of \method. After completing the environment-provided task, the agent uses the remaining interaction budget to explore scene-related goals involving object affordances, state-changing interactions, and interaction compositions. An independent visual checker assesses exploratory outcomes and supplies feedback for continued exploration. Memory management preserves the complete interaction records, organizes experience within scenes, synthesizes knowledge across scenes, and independently reviews candidate rules. Accepted rules form a frozen global memory that guides execution across tasks and variants.}
\label{fig:architecture}
\end{figure}

\method{} supports rapid adaptation by acquiring environment experience, consolidating it into a global memory, and using that memory to guide subsequent behavior.
A high-level agent interacts with the environment through a set
of robot primitives, such as moving to a target pose, adjusting the robot
posture, controlling the gripper, and VLAs.
Each adaptation instance consists of a scene and an environment-provided task, with a shared budget of $B$ interaction turns for task execution and autonomous exploration.
The method comprises three stages, illustrated in Figure~\ref{fig:architecture}.
Exploration collects task-directed and self-directed experience, memory management consolidates the evidence across adaptation scenes, and deployment uses the resulting global memory.

\subsection{Proactive Environment Exploration}
\label{sec:exploration}
\label{sec:initial}
\label{sec:goals}
The agent first attempts the target task $g_0$.
Following some attempts at $b_{\mathrm{init}}$ turns, it retains the task-execution experience and allocates the remaining $B-b_{\mathrm{init}}$ turns to exploring other scene-related goals.
A language-model proposer uses the observed scene and accumulated execution feedback to generate these goals at three levels.

\paragraph{Object affordances.}
The first level explores atomic or local interactions that reveal how individual objects can be used.
Examples include grasping a bowl, opening a cabinet door, or pressing a faucet control.
These interactions provide concrete evidence about the roles and operations supported by objects in the current scene.
The goal at this level is to establish basic object--action relations that later exploration can build upon.

\paragraph{State-changing interactions.}
The second level explores larger but reasonable changes to the scene to understand its functional semantics.
The agent autonomously proposes goals involving storage, placement, opening, closing, or other meaningful state changes.
These interactions provide evidence about how objects and fixtures can be used to establish different scene states.
For example, a Level-2 task may ask the agent to ``place the bowl into the already opened top drawer and verify that the bowl is stably contained inside.''

\paragraph{Interaction compositions.}
The third level explores how multiple verified interactions can be composed to accomplish a more complex goal.
For example, the agent may first pick up a vegetable, rinse it under running water, transfer it into a bowl, and then complete the final placement.
Such goals require the agent to reason about execution order and intermediate conditions connecting the component skills.
% The distinction from the second level is therefore compositional: a long or multi-step behavior remains a state-changing interaction if it transfers as a single reusable unit, while a level-three goal explicitly combines multiple such skills toward a larger task objective.

\paragraph{Visual verification and continued exploration.}
Vision-language models have been studied as success detectors for language-specified tasks \citep{du2023successvqa}.
An independent visual checker assesses each attempted exploration goal using its description and visual observations of the execution outcome.
Its assessment provides feedback for subsequent goal selection and strategy adjustment.
The agent retains the complete interaction records, including successful, failed, and uncertain attempts, together with recovery behavior and visual assessments.
An ordinary failure leads to a change of strategy and further exploration from the current state without resetting the environment.
Both continued attempts and recovery interactions contribute experience within the remaining budget.

\subsection{Memory Management}
\label{sec:memory}
After exploration, \method{} consolidates experience from both the provided task and self-proposed goals.
The process consists of lossless archival, experience atomization, within-scene induction, cross-scene synthesis, independent evidence review.
Its output is a global memory $\mathcal{M}_{\mathrm{global}}$ containing accepted knowledge synthesized across the adaptation scenes.

\paragraph{Lossless archival and experience atomization.}
The complete recorded trajectories are first preserved in an evidence archive $\mathcal{E}$.
This archive retains successes, failures, uncertain outcomes, and recoveries from both experience sources, together with their recorded actions and visual assessments.
Experience atomization then organizes these trajectories into individual experience units for subsequent induction.
Each unit remains associated with its source interaction evidence, allowing later summaries and candidate rules to be reviewed against the original records.

\paragraph{Within-scene induction and cross-scene synthesis.}
Within each scene, the experience units are organized into three forms of knowledge corresponding to the exploration levels.
Object-level knowledge summarizes observed object uses and relations.
Action-level knowledge summarizes state-changing operations and the conditions observed during their execution.
Compositional knowledge summarizes interaction sequences and their intermediate conditions.
Cross-scene synthesis then combines the scene-level summaries into candidate rules about object use, action effects, and interaction composition.
These candidates consolidate experience into knowledge intended to guide different tasks and variants.

\paragraph{Independent evidence review.}
An independent critic reviews the candidate rules against their action evidence and independent visual assessments.
A rule is marked as \texttt{accepted} only when the recorded actions and visual verification support its stated conclusion and it passes the evidence review.
The review considers the support for each conclusion drawn from the archived successes, failures, uncertain outcomes, and recoveries.
Accepted rules are collected into $\mathcal{M}_{\mathrm{global}}$.
The complete evidence archive is retained separately; only accepted knowledge is eligible for injection during evaluation.
Algorithm~\ref{alg:explorer} summarizes exploration and memory consolidation across adaptation instances.

\subsection{Memory-Guided Task Execution}
\label{sec:deployment}
After the memory consolidation stage, the agent combines accepted knowledge from $\mathcal{M}_{\mathrm{global}}$ with the task instruction and current observations.
This shared memory provides guidance on object selection, changed states, and intermediate conditions across different tasks and variants. % 这里对应下游任务的评测结果，突出每层目标对于任务的贡献。例如 goal 就是 obj selection / action skills，而 libero-10 对应 L3。
The agent uses current observations to apply that guidance to the objects, states, and goals of the present task, then executes through robot primitives. At evaluation time, the final global memory $\mathcal{M}_{\mathrm{global}}$ is injected into the agent's context together with the task instruction and current observations. Both the global memory and the VLA parameters remain fixed throughout evaluation.

\begin{algorithm}[t]
    \caption{Proactive exploration and global memory construction with \method}
    \label{alg:explorer}
    \begin{algorithmic}[1]
    \Require instances $\{(e_i,g_i)\}_{i=1}^{N}$, interaction budget $B$ per instance, agent $\mathcal{A}$, robot primitives $\mathcal{P}$
    \State Initialize interaction records $\mathcal{T}\gets\varnothing$
    \For{each instance $(e_i,g_i)$}
        \State Attempt $g_i$ using $\mathcal{A}$ and $\mathcal{P}$
        \State Retain the complete task-execution record in $\mathcal{T}$
        \While{interaction budget remains}
            \State Propose an exploration goal using the three-level objective and accumulated feedback
            \State Execute the goal using $\mathcal{A}$ and $\mathcal{P}$
            \State Assess the outcome with the visual checker
            \State Append the interaction record and assessment to $\mathcal{T}$
            \State Update the exploration strategy from the accumulated feedback
        \EndWhile
    \EndFor
    \State $\mathcal{E}\gets\operatorname{LosslessArchive}(\mathcal{T})$
    \State $\mathcal{U}\gets\operatorname{AtomizeExperience}(\mathcal{E})$
    \For{each adaptation scene $s$ represented in $\mathcal{E}$}
        \State $K_s\gets\operatorname{WithinSceneInduction}(\mathcal{U}_s)$
    \EndFor
    \State $K_{\mathrm{candidate}}\gets\operatorname{CrossSceneSynthesis}(\{K_s\})$
    \State Independently review $K_{\mathrm{candidate}}$ against the action evidence and visual assessments in $\mathcal{E}$
    \State $\mathcal{M}_{\mathrm{global}}\gets
        \operatorname{AcceptedRules}(K_{\mathrm{candidate}})$
    \State \Return $\mathcal{M}_{\mathrm{global}}$
    \end{algorithmic}
\end{algorithm}
\section{Experiments}
\label{sec:experiments}
We evaluate \method{} with one reference instance per task, where the agent is given a fixed interaction budget before being evaluated, testing whether proactively acquired environment knowledge transfers beyond the conditions encountered during adaptation. Harness VLA serves as the task-refinement baseline. Both \method{} and Harness VLA receive the same reference instance.
On RoboCasa365 Composite-Seen, we evaluate the adaptation strategy on longer-horizon household tasks that require multiple interactions to be coordinated within a scene. 
We then study three components of the adaptation process, including the role of the target task, the three-level organization of exploratory goals, and the memory-management process that consolidates exploration evidence into reusable knowledge.

\subsection{Experimental Setup}
\label{sec:setup}
\paragraph{LIBERO-Pro.}
LIBERO-Pro introduces controlled perturbations to the LIBERO manipulation benchmark \citep{liu2023libero,zhou2025liberopro}.
We evaluate Spatial, Object, Goal, and LIBERO-10 under instruction redirection (T) and position swap (S).
Instruction redirection changes the requested target object or goal condition, while position swap changes object placement with the instruction held fixed.
Following the evaluation configuration of Harness VLA~\citep{zhang2026harnessvla},
each of the eight settings contains 10 tasks, and each task is evaluated over 10 rollouts, one for each evaluation seed.
The baselines use their publicly reported results.
The LIBERO-Pro overall score is the unweighted mean over its eight settings.
Cap-X and RATS report only the six non-LIBERO-10 settings, and their marked averages use those reported entries.

\paragraph{RoboCasa365 Composite-Seen.}
We additionally evaluate the 16 Composite-Seen tasks from RoboCasa365 \citep{nasiriany2026robocasa365}.
These tasks span multi-step kitchen activities of varying horizon, ranging from short compositions to routines involving many sequential interactions and intermediate states.
All 16 task classes are represented in the benchmark's pretraining data, while evaluation is performed on target instances.
For both Harness VLA and \method{}, each task is evaluated over five rollouts, one for each evaluation seed, with a budget of 30 agent turns per rollout.
This setting therefore examines whether the acquired memory remains useful for longer-horizon household tasks when the agent must act under a strict evaluation-time interaction budget.

\paragraph{Implementation and comparisons.}
We instantiate \method{} on the Harness VLA execution stack with Codex as the high-level agent \citep{zhang2026harnessvla}.
In this instantiation, \method{} uses the remaining interaction budget after task success for scene-related self-proposed goals, organizes exploration around object affordances, state-changing interactions, and interaction compositions, and verifies exploratory outcomes with an independent visual checker.
The resulting successful, failed, uncertain, and recovery trajectories are processed through lossless archival, experience atomization, within-scene induction, cross-scene synthesis, independent evidence review, and memory freezing.
Harness VLA is evaluated with the same Codex planner, providing a matched task-refinement baseline.

\begin{table}[t]
\caption{
Success rates (\%) on LIBERO-Pro.
T denotes task perturbation and S denotes position swap.
Overall averages use all eight settings except for Cap-X and RATS, whose averages use their six reported settings and are marked with $\dagger$.
A slash indicates an unreported baseline entry.
Bold numbers identify the highest score in each column.
}
\label{tab:main_libero}
\centering
\small
\setlength{\tabcolsep}{2.5pt}
\renewcommand{\arraystretch}{1.08}

\begin{tabular}{@{}p{4.5cm}rrrrrrrrr@{}}
\toprule
\textbf{Method} &
\textbf{Spat-T} &
\textbf{Spat-S} &
\textbf{Obj-T} &
\textbf{Obj-S} &
\textbf{Goal-T} &
\textbf{Goal-S} &
\textbf{L10-T} &
\textbf{L10-S} &
\textbf{Overall} \\
\midrule

OpenVLA \citep{kim2024openvla}
& 0.0 & 0.0 & 0.0 & 0.0 & 0.0 & 0.0 & 0.0 & 0.0 & 0.0 \\

$\pi_0$ \citep{black2024pi0}
& 0.0 & 0.0 & 0.0 & 2.0 & 0.0 & 0.0 & 0.0 & 0.0 & 0.3 \\

$\pi_{0.5}$ \citep{physicalintelligence2025pi05}
& 1.0 & 20.0 & 1.0 & 17.0 & 2.0 & 38.0 & 1.0 & 8.0 & 11.0 \\

MolmoAct \citep{molmoact2025}
& 0.0 & 0.0 & 0.0 & 6.0 & 0.0 & 0.0 & 6.0 & 0.0 & 1.5 \\

NORA \citep{hung2025nora}
& 0.0 & 0.0 & 0.0 & 0.0 & 0.0 & 0.0 & 0.0 & 0.0 & 0.0 \\

X-VLA \citep{zheng2026xvla}
& 0.0 & 0.0 & 8.0 & 2.0 & 9.0 & 1.0 & 10.0 & 0.0 & 3.8 \\

AtomVLA \citep{sun2026atomvla}
& 1.0 & 16.0 & 0.0 & 10.0 & 11.0 & 2.0 & 9.0 & 1.0 & 6.3 \\

\midrule

Cap-X \citep{fu2026capx}
& 14.0 & 12.0 & 18.0 & 22.0 & 17.0 & 26.0 & / & / & 18.2$^{\dagger}$ \\

RATS \citep{zhang2026playful}
& 31.0 & 29.0 & 63.0 & 61.0 & 36.0 & 43.0 & / & / & 43.8$^{\dagger}$ \\

\midrule

$\pi_{\mathrm{RLinf}}$ \citep{yu2026rlinf}
& 42.0 & 59.0 & 71.0 & 78.0 & 45.0 & 42.0 & 49.0 & 14.0 & 50.0 \\

Harness VLA \citep{zhang2026harnessvla}
& 81.0 & 69.0 & 94.0 & 91.0 & 75.0 & 66.0 & \textbf{52.0} & 49.0 & 72.1 \\

\midrule

\textbf{\method{}}
& \textbf{85.0} & \textbf{75.0} & \textbf{98.0} & \textbf{92.0}
& \textbf{88.0} & \textbf{71.0} & 44.0 & \textbf{58.0}
& \textbf{76.4} \\

\bottomrule
\end{tabular}
\end{table}

\begin{table*}[t]
\centering
\caption{
Per-task success rates (\%) on the 16 RoboCasa365 Composite-Seen tasks.
GR00T N1.5~\citep{nvidia_gr00t_n1_5_2025} is evaluated with 30 rollouts per task, while the StarVLA~\citep{community2026starvla} variants
use 50 rollouts per task.
Harness VLA~\citep{zhang2026harnessvla} and \method{} are evaluated on 5 evaluation rollouts per task, one for each evaluation seed.
}
\label{tab:main_robocasa}
\resizebox{\textwidth}{!}{
\begin{tabular}{lcccccc}
\toprule
\textbf{Task}
& GR00T N1.5
& StarVLA-OFT
& StarVLA-PI
& StarVLA-GR00T
& Harness VLA
& \method{} \\
\midrule
DeliverStraw
& 3.0 & 0.0 & 4.0 & 0.0 & \textbf{20.0} & \textbf{20.0} \\

GetToastedBread
& 0.0 & 4.0 & \textbf{56.0} & 52.0 & 20.0 & 20.0 \\

KettleBoiling
& 53.0 & 28.0 & 54.0 & \textbf{60.0} & 20.0 & 20.0 \\

LoadDishwasher
& 47.0 & 30.0 & \textbf{64.0} & 42.0 & 60.0 & 40.0 \\

PackIdenticalLunches
& 17.0 & 2.0 & 0.0 & 0.0 & \textbf{20.0} & \textbf{20.0} \\

PreSoakPan
& \textbf{70.0} & 28.0 & 48.0 & 58.0 & 60.0 & 60.0 \\

PrepareCoffee
& 13.0 & 2.0 & \textbf{28.0} & \textbf{28.0} & 0.0 & 0.0 \\

RinseSinkBasin
& 60.0 & 28.0 & 76.0 & \textbf{80.0} & 20.0 & 40.0 \\

ScrubCuttingBoard
& 70.0 & 8.0 & 54.0 & 18.0 & \textbf{80.0} & \textbf{80.0} \\

SearingMeat
& \textbf{27.0} & 0.0 & 12.0 & 10.0 & 0.0 & 0.0 \\

SetUpCuttingStation
& 33.0 & 10.0 & 30.0 & \textbf{40.0} & 20.0 & \textbf{40.0} \\

StackBowlsCabinet
& 83.0 & 64.0 & 60.0 & 72.0 & \textbf{100.0} & \textbf{100.0} \\

SteamInMicrowave
& 30.0 & 8.0 & 26.0 & 12.0 & 20.0 & \textbf{60.0} \\

StirVegetables
& 33.0 & 0.0 & 10.0 & 24.0 & 0.0 & \textbf{60.0} \\

StoreLeftoversInBowl
& \textbf{43.0} & 18.0 & 40.0 & 4.0 & 0.0 & 0.0 \\

WashLettuce
& 67.0 & 20.0 & 68.0 & 66.0 & 40.0 & \textbf{100.0} \\

\midrule
\textbf{Average}
& 40.6
& 15.6
& 39.4
& 35.4
& 30.0
& \textbf{41.3} \\
\bottomrule
\end{tabular}
}
\end{table*}

\subsection{Overall Results}
\label{sec:overall}
\method{} achieves the highest overall success among the methods in Table~\ref{tab:main_libero}, improving over Harness VLA (Codex) in seven of the eight settings.
This supports the value of using interaction to acquire and consolidate knowledge beyond the requirements of the provided task.

The largest improvement occurs on Goal-T, where execution must satisfy a redirected objective.
All four position-swap settings also improve, showing that the benefit extends to altered spatial arrangements with the instruction held fixed.

On RoboCasa365 Composite-Seen, \method{} outperforms Harness VLA, as shown in Table~\ref{tab:main_robocasa}.
The gains are concentrated in several multi-step tasks, including \texttt{RinseSinkBasin}, \texttt{SetUpCuttingStation}, \texttt{SteamInMicrowave}, \texttt{StirVegetables}, and \texttt{WashLettuce}.
These tasks require the agent to coordinate multiple interactions and intermediate states within a limited number of decisions.
The result suggests that proactively acquired environment knowledge can make the available evaluation budget more useful by providing reusable information about object roles, state changes, and interaction sequences before evaluation begins.

\subsection{Ablations}
\label{sec:ablations}

Table~\ref{tab:ablations} isolates three components of \method{} on LIBERO-Pro Goal-S.
\textbf{w/o initial task guidance} skips the environment-provided task and allocates the adaptation budget directly to autonomous exploration.
\textbf{w/o memory management} keeps the collected interaction experience but bypasses the three-level memory consolidation, using the raw exploration memory for execution.
\textbf{w/o three-level exploration} retains the same interaction budget and memory-management pipeline, but removes the three-level constraint on goal generation and instead allows free-form scene exploration.

Removing initial task guidance causes only a mild decrease, suggesting that autonomous exploration alone can already acquire useful environment experience, while the provided task gives exploration an additional task-relevant starting point.
Removing memory management produces a similarly moderate reduction, indicating that organizing raw interaction records into reusable memory improves how the acquired experience can be used during task execution.

The largest degradation occurs without three-level exploration.
This result indicates that the content of exploration matters more than simply increasing the amount of interaction.
Free-form exploration still provides environment experience, but organizing goals around object affordances, state-changing interactions, and interaction compositions produces substantially more useful knowledge for later task execution.

\begin{table}[t]
\caption{Task-level success rates (\%) on the LIBERO-Pro Goal-S setting.
Task IDs and their corresponding instructions are listed in Appendix~\ref{app:goal_s_tasks}.}
\label{tab:ablations}
\centering
\small
\setlength{\tabcolsep}{5pt}
\begin{tabular}{@{}lcccc@{}}
\toprule
\textbf{Task} &
\textbf{\method{}} &
\makecell{\textbf{w/o initial}\\\textbf{task guidance}} &
\makecell{\textbf{w/o memory}\\\textbf{management}} &
\makecell{\textbf{w/o three-level}\\\textbf{exploration}} \\
\midrule
task00 & 0.0   & 0.0   & 0.0   & 0.0   \\
task01 & 90.0  & 90.0  & 80.0  & 30.0  \\
task02 & 70.0  & 80.0  & 80.0  & 50.0  \\
task03 & 80.0  & 70.0  & 80.0  & 10.0  \\
task04 & 100.0 & 100.0 & 100.0 & 90.0  \\
task05 & 20.0  & 0.0   & 10.0  & 0.0   \\
task06 & 70.0  & 70.0  & 80.0  & 30.0  \\
task07 & 100.0 & 100.0 & 100.0 & 100.0 \\
task08 & 100.0 & 100.0 & 90.0  & 70.0  \\
task09 & 80.0  & 70.0  & 50.0  & 40.0  \\
\midrule
Overall & \textbf{71.0} & 68.0 & 67.0 & 42.0 \\
\bottomrule
\end{tabular}
\end{table}

\section{Analysis}
\label{sec:analysis}

% We analyze how experience acquired during adaptation is reflected in subsequent execution and stored knowledge.
% The study connects success under limited VLA invocations with the agent's action instructions, tool transitions, and use of unsuccessful experience.

\begin{figure}[t]
    \centering
    \begin{minipage}[t]{0.49\linewidth}
        \centering
        \includegraphics[width=\linewidth]{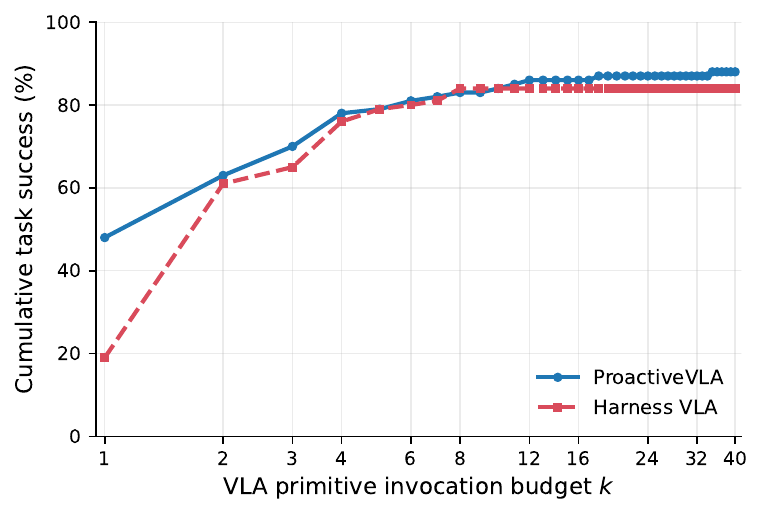}
        {\small (a) Goal-T comparison}
    \end{minipage}\hfill
    \begin{minipage}[t]{0.49\linewidth}
        \centering
        \includegraphics[width=\linewidth]{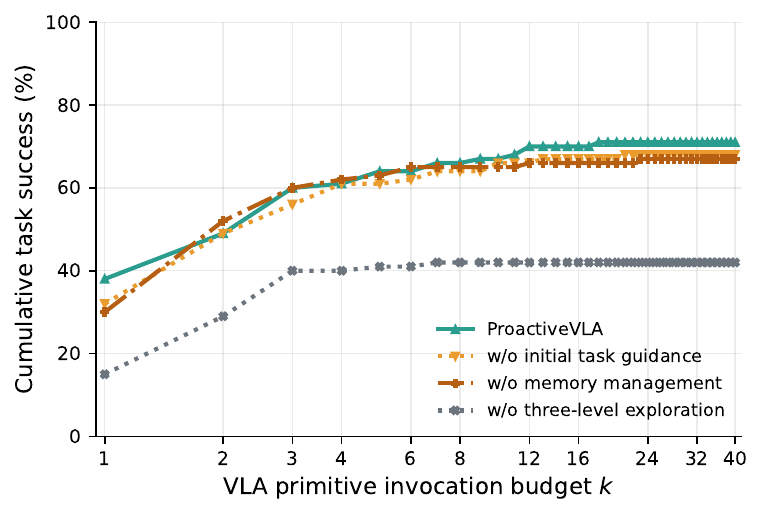}
        {\small (b) Goal-S component ablations}
    \end{minipage}
    \caption{Task success as a function of the maximum number of VLA primitive invocations allowed during evaluation.
    \textbf{(a)} \method{} has its largest advantage over Harness VLA when only a few VLA calls are available.
    \textbf{(b)} Structured three-level exploration contributes most strongly to this low-budget execution capability.}
    \label{fig:budget_curves}
\end{figure}

\paragraph{Proactive exploration improves execution efficiency under limited VLA calls.}
\label{sec:analysis_budget}
Figure~\ref{fig:budget_curves} reports the fraction of evaluation instances completed with at most $k$ VLA primitive invocations.
Figure~\ref{fig:budget_curves}(a) reports our Goal-T re-evaluation for the VLA-invocation analysis. \method{} completes 48\% of instances with at most one VLA call, compared with 19\% for Harness VLA.
The strongest separation therefore appears at the smallest call budget, where additional controller calls are unavailable to compensate for a less effective execution strategy.
This result supports the value of proactively acquired experience for completing tasks with limited use of the frozen controller, beyond improving eventual success alone.
Here, $k$ counts VLA calls during evaluation, while perception and analytic control tools remain available before and after each call.
Figure~\ref{fig:budget_curves}(b) further shows that this behavior depends on how the experience is acquired and organized. The much larger drop without three-level exploration shows that free interaction alone is insufficient.
Organizing exploration around object affordances, state-changing interactions, and interaction compositions is important for turning a limited adaptation opportunity into knowledge that remains useful during task execution.

\begin{figure}[t]
    \centering
    {\small\bfseries (a) \method{}\par}
    \smallskip
    \includegraphics[page=1,width=\linewidth]{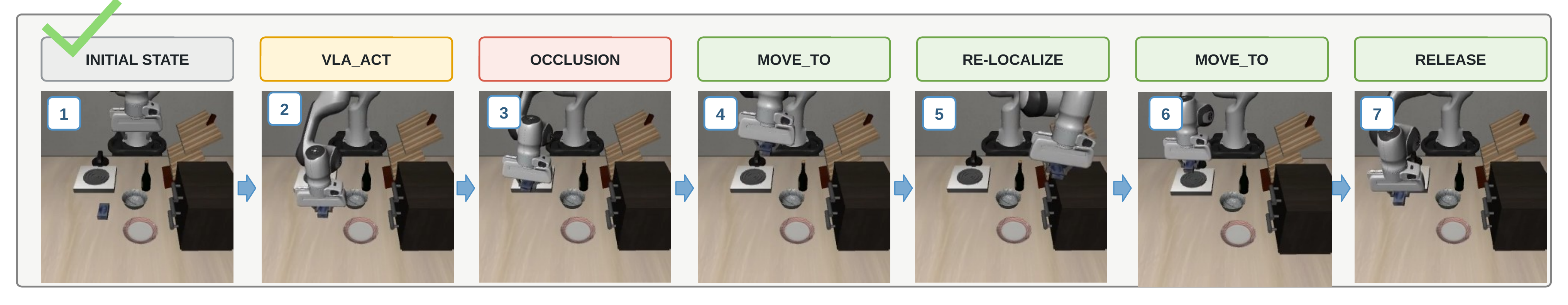}
    \medskip
    {\small\bfseries (b) Harness VLA\par}
    \smallskip
    \includegraphics[page=2,width=\linewidth]{figures/occlusion_recovery.pdf}
    \caption{Execution under target occlusion.
    Both agents move the arm after the target becomes occluded.
    \method{} re-localizes before resuming the task and succeeds, while Harness VLA skips re-localization, acts at the wrong position, and fails.}
    \label{fig:occlusion_case}
\end{figure}

\paragraph{Exploration-derived recovery rules guide execution when conditions change.}
Proactive exploration also produces conditional rules that can be reused when execution departs from the nominal trajectory.
Figure~\ref{fig:occlusion_case} illustrates this behavior under target occlusion.
After the manipulator blocks the object required for localization, both agents first move the arm away.
\method{} then re-localizes the target before resuming execution and successfully completes the task.
Harness VLA skips the re-localization step, continues operating at an incorrect location, and eventually fails.
The global memory acquired by \method{} contains the corresponding rule that when the manipulator occludes an object needed for localization, the agent should first move away and then localize the object again before continuing.

\paragraph{Proactive exploration leads to finer-grained use of the VLA controller.}
Harness VLA more often gives the VLA task-level instructions such as ``open the bottom drawer of the cabinet,'' whereas \method{} more often uses action-level instructions that specify the contact target, motion direction, and stopping condition, such as ``grasp the lowest drawer's gray handle, then pull it straight outward until the bottom drawer is clearly open.''
Action-level instructions account for 63.4\% of \method{}'s VLA calls versus 34.7\% for Harness VLA, while task-level instructions account for only 0.6\% versus 27.3\%. The difference is also reflected in tool use.
After a VLA call, \method{} transitions to \texttt{move\_to} in 24.7\% of cases, compared with 8.4\% for Harness VLA, while immediate VLA-to-VLA continuations are less frequent.
These results indicate that proactive exploration helps the agent decompose execution into more specific operations and coordinate the VLA with other control tools.
\section{Limitations}

Our experiments are conducted in simulated manipulation environments, and the extent to which the acquired memory transfers to real-world perception, dynamics, and execution noise remains to be evaluated.
Although \method{} is formulated around a general agent--primitive interface, our current implementation is instantiated on a single agentic VLA execution stack, so its effectiveness with different planners and low-level controllers is not yet established.
The evaluation also uses fixed interaction budgets; different amounts or distributions of adaptation experience may change the relative benefit of proactive exploration and task refinement.
Future work should evaluate the same experience-acquisition strategy across real robots, alternative system architectures, and a wider range of adaptation budgets.
\section{Conclusion}
\label{sec:conclusion}
We study how agentic VLA systems can use limited interaction with a new environment to acquire reusable knowledge for rapid adaptation.
\method{} uses the remaining interaction budget after task completion for structured proactive exploration and consolidates successful, failed, and recovery experience into a global memory while keeping the VLA frozen.
Experiments on LIBERO-Pro and RoboCasa365 Composite-Seen show that this strategy improves task execution over continued task refinement, with particularly benefits when only a few VLA calls are available.
These results suggest that effective adaptation depends not only on accumulating experience, but on actively acquiring and organizing environment knowledge that can guide subsequent behavior.
\section*{AI Use Statement}

Generative AI tools were used throughout several stages of this work. During research ideation and execution, we used AI tools to brainstorm and refine research ideas, assist with experimental design and implementation, and support the analysis of intermediate experimental results. We also used generative AI to produce synthetic data used in our experiments, including the exploration task and memory generation described in the paper. For literature research, AI tools were used to assist with retrieval and discovery of relevant work and to summarize candidate literature during the search process. 

During manuscript preparation, generative AI tools were used both to draft portions of the paper and to revise existing text for clarity, organization, grammar, and readability. The authors reviewed, revised, and verified all AI-assisted text, experimental content, references, and generated data before inclusion in the final manuscript. The authors take full responsibility for the content, claims, results, and conclusions of this work.
\bibliographystyle{iclr2027_conference}
\bibliography{references}
% \clearpage
\appendix
\section{LIBERO-Pro Goal-S Tasks}
\label{app:goal_s_tasks}

Table~\ref{tab:goal_s_tasks} lists the 10 tasks used in the LIBERO-Pro Goal-S evaluation and their correspondence to the task IDs reported in Table~\ref{tab:ablations}.

\begin{table}[h]
\caption{Tasks in the LIBERO-Pro Goal-S evaluation.}
\label{tab:goal_s_tasks}
\centering
\small
\begin{tabular}{@{}ll@{}}
\toprule
\textbf{Task ID} & \textbf{Task instruction} \\
\midrule
task00 & Open the middle drawer of the cabinet \\
task01 & Put the bowl on the stove \\
task02 & Put the wine bottle on top of the cabinet \\
task03 & Open the top drawer and put the bowl inside \\
task04 & Put the bowl on top of the cabinet \\
task05 & Push the plate to the front of the stove \\
task06 & Put the cream cheese in the bowl \\
task07 & Turn on the stove \\
task08 & Put the bowl on the plate \\
task09 & Put the wine bottle on the rack \\
\bottomrule
\end{tabular}
\end{table}

\section{Prompts Used by \method{}}
\label{app:prompts}

\subsection{Proactive Environment Exploration}
\label{app:prompt_exploration}

\paragraph{Provided-task acquisition prompt.}
\begin{quote}\small
During the task-guided acquisition stage, focus only on the provided task:
\texttt{\{task\_language\}}.

Prioritize completing the task and extracting transferable interaction patterns.
The environment's official success signal is the only success criterion for the provided task.
Do not autonomously propose additional tasks during this stage.

Budget: at most \texttt{\{max\_turns\}} model requests and
\texttt{\{max\_actions\}} environment actions.
\end{quote}

\paragraph{Proactive exploration prompt.}
\begin{quote}\small
You are not given an external task and must not read, infer, or reconstruct hidden benchmark goals.
Your objective is to continue understanding the same scene and acquire high-quality knowledge that can transfer to future tasks.
Proceed through the following three levels in order.

\textbf{Level 1, object affordances.}
Prioritize testing functional objects such as cabinets, drawers, and knobs for opening, closing, feedback, and internal contents.
Then test how different objects can be grasped, moved, and returned.

\textbf{Level 2, state-changing interactions.}
Perform larger but reasonable state changes to understand the semantics of the scene.
Autonomously propose interaction goals involving operations such as storage, placement, opening, closing, or other meaningful state changes.

\textbf{Level 3, interaction compositions.}
Construct longer tasks only by composing skills that have already been verified.
For example, combine object placement with a subsequent fixture interaction.
Distinguish support surfaces from container interiors.

\textbf{Task-design principles.}
\begin{enumerate}
    \item A Level-1 task must be atomic and contain exactly one visually verifiable subgoal, such as opening one drawer and confirming its displacement, or lifting one object and confirming that it has left its support surface. Do not combine grasping, transporting, and returning an object into a single Level-1 task.
    \item Prefer goals whose before-and-after differences are clear, unobstructed, and directly visible from the global view. After an action, first move to an unobstructed observation pose and wait for the image to stabilize before judging the outcome.
    \item If the evidence clearly shows that the goal was not completed, record the result as failure. Use uncertain only when the visual evidence is genuinely insufficient.
    \item Generate memory from an autonomous task only after \texttt{complete\_task} returns \texttt{verification.independently\_verified=true}. Otherwise, continue with a simpler and more falsifiable task.
    \item The success criterion for an autonomous task must rely only on task-local visual evidence or observable state change. Do not use \texttt{terminated}, \texttt{truncated}, official success, benchmark signals, or hidden task information during exploration.
\end{enumerate}

\textbf{Hard constraints.}
\begin{enumerate}
    \item Call \texttt{propose\_task} before the first action of every autonomous task, and work on only one task at a time.
    \item At the end of every task, call \texttt{complete\_task} and then \texttt{add\_memory\_candidate}. Record failures and uncertain outcomes as well as successes.
    \item Memory must use semantic or relative descriptions. Include applicability conditions, verification evidence, failure modes, and recovery strategies. Do not store historical absolute coordinates.
    \item Do not reset by default. Request an audited reset only when an irreversible error would prevent further useful exploration. Ordinary failures must not trigger a reset.
    \item Complete at least three Level-1 tasks, two Level-2 tasks, and one Level-3 task, and produce at least three controller-verified memory candidates before calling \texttt{finish\_exploration}.
    \item Self-evaluation in text is not evidence. Cite the actual action call IDs and preserve the continuous state of the same scene while observing before-and-after changes.
\end{enumerate}

Total budget: at most \texttt{\{max\_turns\}} model requests and
\texttt{\{max\_actions\}} environment actions.
\end{quote}

\paragraph{Exploration retry policy.}
\begin{quote}\small
For an autonomous task, do not abandon the task or switch to a new one while it remains unverified and both the task-level and global action budgets remain available.
Reserve at least three environment-action attempts for each task.

After a failed grasp, opening, closing, or rotation attempt, continue within the same task.
Use the observed failure evidence to change the approach direction, grasp point, gripper pose, or action magnitude.
Do not repeat an identical strategy without new information.

An occluded view cannot be treated as either success or final failure.

Every tool result contains an explicit \texttt{ledger\_call\_id}.
When calling \texttt{complete\_task}, \texttt{evidence\_calls} must reference the calls that actually executed the relevant action and the final unobstructed observation.
Do not cite only segmentation, localization, or an earlier approach action.

The \texttt{ledger\_call\_id} is only an evidence identifier and is not an environment step.
For \texttt{segment} or \texttt{back\_project}, use the most recent \texttt{result.step}.
If the current step is uncertain, use \texttt{step=-1} to retrieve the latest state.

Finalize a task only when it has been independently verified as successful, when its action budget is exhausted, or when recovery attempts confirm an irreversible error.
\end{quote}

\paragraph{Visual verification prompt.}
\begin{quote}\small
You are a visual evidence verifier completely isolated from the exploration planner.
You do not know the benchmark goal and must not infer it.
Verify only whether the autonomous task proposal and completion claim are directly supported by the provided pre-action image, relevant action results, and post-action image.

The verdict concerns whether the autonomous task itself was completed, not whether the textual claim is honest.
If the claim correctly states that the task failed and the images confirm failure, the task verdict must still be \texttt{failed}.
The overall verdict must agree with the subgoal verdicts.
All subgoals must be \texttt{passed} for the overall verdict to be \texttt{passed}.
If any subgoal is \texttt{failed}, the overall verdict is \texttt{failed}.
Otherwise, use \texttt{uncertain}.

\textbf{Decision rules.}
\begin{enumerate}
    \item \texttt{passed}: every claimed subgoal is directly confirmed by a clear before-and-after visual change.
    \item \texttt{failed}: the images or action results directly contradict the claim, or clearly show that at least one claimed subgoal was not completed.
    \item \texttt{uncertain}: evidence is missing, the view is occluded, the target cannot be identified, only robot contact is visible without a corresponding object or fixture state change, or the available evidence cannot establish the outcome.
    \item Tool-returned \texttt{completed}/\texttt{success}, physical contact, or execution of a trajectory is not sufficient visual evidence of success.
    \item If a task claims that a knob was turned and a stove was activated, but the evidence only shows robot contact and no visually verifiable knob or stove change, the verdict must be \texttt{failed} or \texttt{uncertain}, never \texttt{passed}.
    \item For each subgoal, return the verdict, localized supporting evidence, contradictions, and confidence. Do not introduce information absent from the supplied evidence.
\end{enumerate}

You must call \texttt{submit\_visual\_verdict} to return the structured result.
\end{quote}

\paragraph{Provided-task memory distillation prompt.}
\begin{quote}\small
The provided task has been confirmed successful by the environment's official success criterion.
Distill the trajectory into one semantic memory item that can transfer across seeds.
Describe object-relative relations, the action skeleton, success checks, common failure modes, and recovery strategies.
Do not store world coordinates or pixel coordinates, and do not merely restate the task.

Task: \texttt{\{task\_language\}}

Evidence calls: \texttt{\{evidence\}}
\end{quote}

\subsection{Memory Management}
\label{app:prompt_memory}

\paragraph{Memory-consolidation system prompt.}
\begin{quote}\small
You are an experience induction module for a robot-learning experiment.
Use only the provided seed-0 evidence.
Separate observed facts from causal hypotheses.
Do not use evaluation tasks, hidden rewards, or privileged simulator state.
Output only semantic experience that can transfer across seeds.
Do not use absolute world coordinates or pixel coordinates.
Return the result through the specified function.
\end{quote}

\paragraph{Within-scene three-level induction prompt.}
\begin{quote}\small
Organize all attempts from this scene, including failures, retries, and recoveries, into three levels of candidate notes.

\begin{enumerate}
    \item \texttt{object\_affordance}: object properties, possible interactions, and object-conditioned operations.
    \item \texttt{action\_skill}: observation, approach, grasping, movement, placement, pushing, pulling, rotation, occlusion removal, and recovery.
    \item \texttt{long\_horizon}: step transitions, interference control, checkpoints, and recovery branches.
\end{enumerate}

Both successes and failures may provide evidence.
When the cause is uncertain, record it in \texttt{counterexamples} and reduce confidence.
Do not turn a single observation into an unconditional rule.
Every \texttt{evidence\_refs} field must copy an \texttt{evidence\_ref} that exists in the input.
Do not restate the task and do not write coordinates.

Method: \method{}

Seed-0 task: \texttt{\{task\}}

Evidence: \texttt{\{evidence\}}
\end{quote}

\paragraph{Cross-scene synthesis prompt.}
\begin{quote}\small
Deduplicate and synthesize the notes from the ten scenes into three levels of cross-seed candidate knowledge.
Preserve applicability conditions, failure counterexamples, and recovery methods.
Increase confidence only when a statement is supported by at least two independent pieces of evidence or by a success-failure contrast.
A conclusion supported by a single piece of evidence must retain an explicitly limited scope.
Do not add facts that are absent from the scene notes.
Do not use coordinates.
All \texttt{evidence\_refs} must preserve existing references and must not be invented.

Method: \method{}

Scene notes: \texttt{\{scene\_notes\}}
\end{quote}

\paragraph{Independent evidence-review prompt.}
\begin{quote}\small
You are an independent evidence reviewer.
For each candidate, verify whether its stated conclusion is directly supported by the cited original evidence.
The review asks whether the candidate is suitable as conditional experimental memory formed only from seed-0 interaction.
It does not require the candidate to have already demonstrated a global causal effect on evaluation seeds.

A single independently verified success or failure may support a narrowly scoped observation, checking rule, or same-class task step, but it cannot support a universal causal advantage.

Use the following verdicts.

\textbf{\texttt{accepted}}:
the core content is directly supported by evidence, the applicability range is reasonable, single-sample limitations are explicit where needed, no coordinates are included, and no privileged information is leaked.

\textbf{\texttt{hypothesis}}:
the statement is potentially useful but the evidence is insufficient, or the proposed cause remains speculative.

\textbf{\texttt{rejected}}:
the candidate contradicts the evidence, has no valid citation, depends on absolute coordinates, or is not reusable.

Do not accept a candidate simply because its wording appears plausible.
\texttt{supported\_evidence\_refs} may contain only keys present in the supplied evidence index.
\texttt{unsupported\_claims} may list only claims actually present in the candidate and unsupported by evidence.
If \texttt{verdict=accepted}, \texttt{unsupported\_claims} must be empty.

Method: \method{}

Candidates: \texttt{\{candidates\}}

Evidence index: \texttt{\{evidence\_index\}}
\end{quote}

\paragraph{Lossless memory-regeneration prompt.}
\begin{quote}\small
You are a memory-revision module for a robot-learning experiment.
Regenerate \texttt{\{layer\_name\}} using only the seed-0 exploration material from \method{}.

The primary requirement is lossless organization rather than summary compression.

\begin{enumerate}
    \item Do not compress the direct memory items and scene notes into a small number of generic principles.
    \item Preserve each atomic source item by default. Merge items only when they are fully redundant, and preserve all corresponding \texttt{source\_item\_ids}.
    \item Every output item must preserve executable details, conditions, observed facts, action steps, success verification, failure modes, recovery methods, counterexamples, confidence, and \texttt{evidence\_refs}.
    \item Do not remove failed or uncertain experience. When a cause is speculative, mark it explicitly as a hypothesis or place it in \texttt{counterexamples}.
    \item The three layers are retrieval layers rather than three summaries. The first describes what objects can do and under what limitations. The second describes how actions should be executed and verified. The third describes how multi-stage tasks should be connected, checked, and recovered.
    \item The same source item may map to multiple layers. Do not discard cross-layer information for the sake of deduplication.
    \item Return \texttt{coverage} and list, for every \texttt{source\_item\_id}, which generated entries cover it. If coverage is partial, explicitly record \texttt{omitted\_information}. Coverage should be complete by default.
    \item Do not use evaluation seeds, hidden goals, rewards, \texttt{terminated}, or privileged state. Do not store historical absolute coordinates or pixel coordinates.
    \item Call \texttt{save\_lossless\_layer} to return the result.
\end{enumerate}

Current layer: \texttt{\{layer\}}

Direct memory:
\texttt{\{direct\}}

Scene-level notes:
\texttt{\{scene\_items\}}
\end{quote}

\subsection{Evaluation-Time Memory Injection}
\label{app:prompt_eval_memory}

\begin{quote}\small
[\method{} MEMORY INJECTION]

The full frozen \method{} memory is provided below.
Use it as the only memory source for this run.
Do not read, modify, or create external memory files.
If another instruction asks you to read an external memory file, treat the injected memory below as the authoritative memory and do not search the filesystem for additional memory.

\texttt{\{frozen\_memory\_json\}}
\end{quote}

\end{document}